\pdfoutput=1

\documentclass[11pt]{article}

\usepackage{ACL2023}

\usepackage{times}
\usepackage{latexsym}
\usepackage{amsmath}
\usepackage{mathrsfs}
\usepackage{tcolorbox} 
\usepackage{booktabs}
\usepackage{multirow}
\usepackage{makecell}
\usepackage{adjustbox}
\usepackage{float}
\usepackage{stfloats}
\usepackage{subcaption}

\tcbuselibrary{skins} 
\usepackage[T1]{fontenc}

\usepackage[T1]{fontenc}

\usepackage[utf8]{inputenc}

\usepackage{microtype}

\usepackage{inconsolata}
\usepackage{graphicx}

\title{LPNL: Scalable Link Prediction with Large Language Models}

\author{Baolong Bi$^{1,3}$ \quad Shenghua Liu$^{1,3}$\thanks{ \ \ Corresponding author.}\quad  Yiwei Wang$^2$ \quad Lingrui Mei$^{1,3}$ \quad Xueqi Cheng$^{1,3}$ \\
	$^1$CAS Key Laboratory of AI Security, Institute of Computing Technology, Chinese Academy of Sciences \\
	$^2$University of California, Los Angeles \\
        $^3$University of Chinese Academy of Sciences \\
	\texttt{\small{\{bibaolong23z,liushenghua,cxq\}@ict.ac.cn}}
	\texttt{  \small{wangyw.evan@gmail.com}}
        \texttt{ \small{meilingrui22@mails.ucas.ac.cn}}
 }

\begin{document}
\maketitle
\begin{abstract}
Exploring the application of large language models (LLMs) to graph learning is an emerging endeavor. 
However, the vast amount of information inherent in large graphs poses significant challenges to graph learning with LLMs. 
This work focuses on the link prediction task and introduces \textbf{LPNL} (\textbf{L}ink \textbf{P}rediction via \textbf{N}atural \textbf{L}anguage), a framework based on large language models designed for scalable link prediction on large-scale heterogeneous graphs.
We design novel prompts for link prediction that articulate graph details in natural language. 
We propose a two-stage sampling pipeline to extract crucial information from the graphs, and a divide-and-conquer strategy to control the input tokens within predefined limits, addressing the challenge of overwhelming information. 
We fine-tune a T5 model based on our self-supervised learning designed for link prediction.
Extensive experimental results demonstrate that LPNL outperforms multiple advanced baselines in link prediction tasks on large-scale graphs.
\end{abstract}

\section{Introduction}

\begin{figure}[t]
    \centering
    \includegraphics[width=0.36\textwidth]{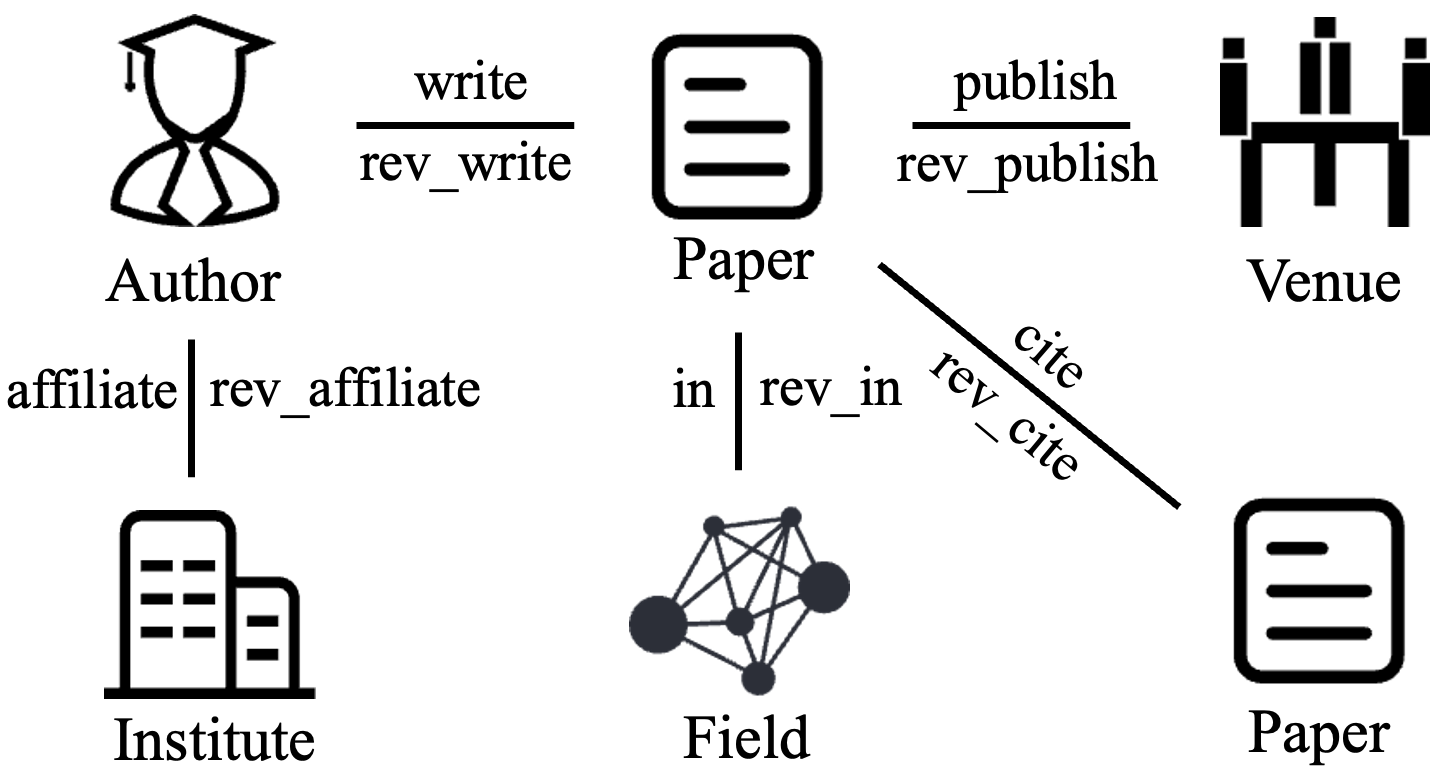}
    \caption{An example of heterogeneous graph}
    \label{fig:fig1}
    \vspace{-5mm}
\end{figure}

Heterogeneous graphs~\cite{shi2016survey} are commonly employed for modeling complex systems, wherein entities of diverse types interact with each other via various relations. 
Figure \ref{fig:fig1} shows the heterogeneous nodes and their relationships sourced from the Open Academic Graph (OAG)~\cite{huang2020analysis}.
Link prediction~\cite{zhang2018link, cai2021line} is a fundamental task in graph learning. 
However, due to the vast quantity of nodes and edges with their complex structure, addressing the link prediction task on large-scale heterogeneous graphs is challenging.

Recently, some research~\cite{fatemi2023talk, ye2023natural} has explored the use of large language models (LLMs) in graph learning. 
A popular paradigm of link prediction on graphs with LLMs is to transform graph problems and structures into description texts, and then feed the texts to LLMs to obtain the predictions. 
However, it remains under explored that how to perform scalable link prediction on large graphs through LLMs with the input window constraints, which poses serious challenges in capturing distant information and rich semantics.  
As the number of nodes increases, the text fed into LLMs grows. Consequently, extensive inputs become unfeasible due to token length limitations.



In this work, we explore the scalable link prediction with large language models on large-scale heterogeneous graphs. The key challenges can be described as follows: 1) how to fomulate the prompt template for scalable link prediction task. 2) how to find out crucial information on large graphs, enabling LLMs to capture it within limited inputs. 3) how to address lengthy prompts generated by an excess of candidate neighbors. To tackle the above challenges, we propose \textbf{LPNL} (\textbf{L}ink \textbf{P}rediction via \textbf{N}atural \textbf{L}anguage), a large language model based framework for scalable link prediction on large-scale graphs.

We design novel prompts for link prediction that articulating graph details in natural language. This involves establishing a selective query prompt template, furnishing a description of the link prediction task, and integrating heterogeneous information concerning the source node and candidate neighbors.

In dealing with vast amounts of relevant graph information within large graphs, LPNL selects crucial node information from the graph, ensuring that the model focuses more on them. We design a two-stage sampling pipeline that utilizes normalized degree-based heterogeneous subgraph sampling and personalized pagerank-based ranking. This approach avoids the interference of superfluous contextual information while ensuring compliance with specified token limitations.

\begin{figure*}[t]
  \centering
  \includegraphics[width=0.95\textwidth]{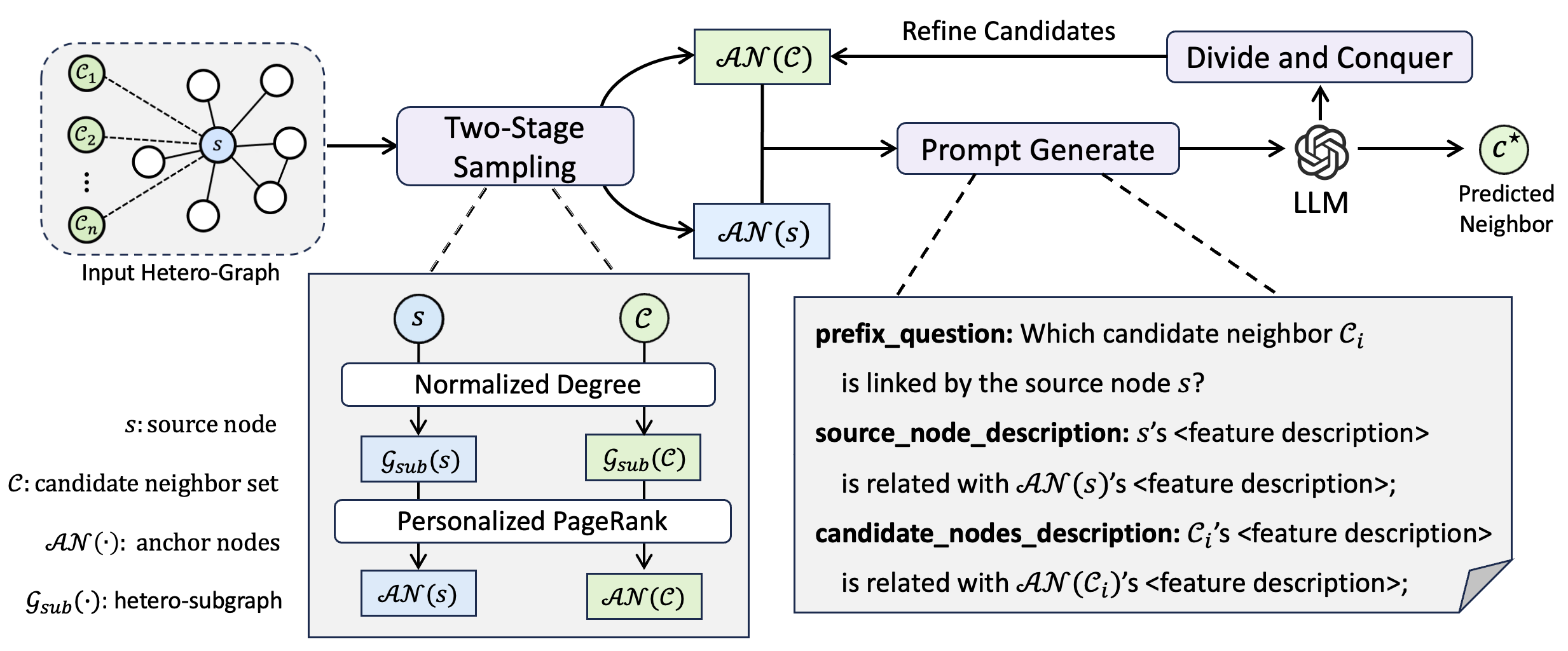}
  \caption{The framework of LPNL. For an input heterogeneous graph with link prediction tasks, LPNL consists of three steps: (1) conduct a two-stage sampling on the source node and each candidate neighbor from the original candidate set to acquire anchor nodes. (2) Generate prompts based on these anchor nodes and input them into LLMs for predictions. (3) Refine the candidate set based on prediction results and iteratively apply this divide-and-conquer process to obtain the distinct link prediction result
  $c^*$.}
  \label{fig:fig2}
\end{figure*}

With a large number of candidate neighbors, the token length constraints make it challenging to fully describe all candidate neighbor information.
To address this issue, we employ a divide-and-conquer method. The original node set is partitioned into multiple sets with smaller size, which are sequentially input into the link prediction pipeline to obtain partial answers. 
Subsequently, we recursively refine the candidate set to predict the final answer.

We conduct extensive experiments on the OAG and fine-tune the language model T5~\cite{raffel2020exploring} based on our self-supervised learning to serve as the backbone model for LPNL on the OAG . 
The results demonstrate that LPNL significantly outperforms various enhanced GNN-based baselines, achieving an average improvement of 30.52\% on Hits@1. 
Furthermore, through extensive experimentation, LPNL also exhibits remarkable few-shot capability. 
Unlike traditional models training, LPNL's fine-tuning merely requires simple alignment formatting, enabling swift convergence in predictions. 
Additionally, experiments demonstrate the model's robust knowledge transferability, maintaining consistent performance across various cross-domain tasks. 
This further emphasizes that LPNL's self-supervised fine-tuning is not confined to fixed graph labels, it can make direct predictions on different graphs without the need for additional learning.

\section{The LPNL Architecture}
In this section, we introduce the details of our proposed \textbf{L}ink \textbf{P}rediction via \textbf{N}atural \textbf{L}anguage, i.e. \textbf{LPNL}, a framework utilizing natural language to solve link prediction task on large-scale 
heterogeneous graphs. 
We start with the notation setup, followed by the prompt design, the sampling methods, and our divide-and-conquer and self-supervised strategy with more details.

\subsection{Preliminary}
Formally, a heterogeneous graph is denoted by $\mathcal{G} = \{\mathcal{V}, \mathcal{E}, \mathcal{A}, \mathcal{R}\} $, where $\mathcal{V}$ and $\mathcal{E}$ denote the sets of nodes and edges (links), respectively. 
Each node $v\in\mathcal{V}$ and each link $e\in\mathcal{E}$ are associated with their mapping function $\phi(v):v\rightarrow \mathcal{A}$ and $\varphi(e):e\rightarrow \mathcal{R}$. 
$\mathcal{A}$ represents the set of node types, and $\mathcal{R}$ represents the set of edge types. 

Given a source node $s$ and a set of candidate neighbors $\mathcal{C}=\{c_1, c_2, ..., c_n\}$, satisfying a existed meta-relation $\langle \phi(s), \varphi(e), \phi(c_i) \rangle$ where $e \in \mathcal{E}$ and $c_i \in \mathcal{C}$, a standard link prediction task on heterogeneous graphs aims to predict a candidate neighbor $c \in \mathcal{C}$ for a source node $s$ with the highest probability of $\langle s, e, c\rangle$.

Finally, let $\mathcal{G}_{sub}^h(v) = \{\mathcal{V}_v^h, \mathcal{E}_v^h, \mathcal{A}_v^h, \mathcal{R}_v^h\}$ denote the $h$-hop ego-subgraph around $v$, consisting of $h$-hop neighbor nodes of $v$ and all interconnecting edges.
We also denote $\mathcal{N}^h(v)$ as the set of all neighbor nodes on $\mathcal{G}_{sub}^h(v)$, which means $\mathcal{N}^h(v) = \{v'\vert v' \in \mathcal{V}_v^h,v'\neq v\}$.
Additionally, $\mathcal{AN}^h_k(v)$ is denoted as the sequence of top-$k$ anchor nodes selected from $\mathcal{N}^h(v)$. Note that all the above definitions are heterogeneous.

\subsection{Prompt Design for Link Prediction}\label{subsec:prompt}
In order to comprehensively represent the link prediction task along with the essential graph information, we meticulously design a uniform prompt template $\mathcal{T}(\cdot)$ for heterogeneous link prediction.
Its fundamental mode involves a selective query, providing both the link prediction problem description and information regarding the source node and candidate neighbors. This prompts the large language models to identify the node most likely to be linked within the candidate set.

First, we define $d(v)$ as the description of node $v$, which consists of a sequence of textual features of itself and also its top-$k$ anchor nodes:
\begin{equation}\label{eq:eq1}
    d(v) = \{v: \mathcal{S}_v\} \text{ is related with } \sum_{i=1}^k\{v'_i:\mathcal{S}_{v'_i}\}
\end{equation}
where $\mathcal{S}_v$ denotes the textual description of node $v$ and $v'_i$ represents the anchor node of node $v$ satisfying $v'_i \in \mathcal{AN}^h_k(v)$. 

Subsequently, given a source node $s$ and the set of candidate neighbors $\mathcal{C}$, we formally obtain the link prediction prompt template as follows:
\begin{equation}\label{eq:eq2}
    \mathcal{T}(s, \mathcal{R}, \mathcal{C}) = q(\mathcal{R}) + d(s) + \sum_{i=1}^n d(c_i\vert c_i \in \mathcal{C})
\end{equation}
where $\mathcal{R}$ is the relation type between the source node and  candidate neighbors and $n$ is the number of candidate neighbors.
And $q(\mathcal{R})$ represents a link prediction query, e.g., "which $\phi(c)$ is linked by $\phi(u)$?". Notably, in the above equation, the addition operators are redefined as the textual concatenation with separators. 

\begin{figure}[ht]
    \centering
\begin{tcolorbox}[fonttitle=\bfseries, title=Author Disambiguation Example, size = normal, label=mybox, center title]
\textbf{prefix\_question}: Which following candidate author writes the paper p$_{1}$?   
\tcbline

\textbf{source\_node\_description}: p$_{1}$: <\textit{paper title}> is related with f$_{25}$: <\textit{field name}>, v$_{13}$: <\textit{journal info}>, p$_{46}$: <\textit{paper title}>, a$_{38}$: <\textit{author info}>, p$_{27}$: <\textit{paper title}>...
\tcbline

\textbf{candidate\_nodes\_description}: a$_{1}$: <\textit{author info}> is related with p$_{15}$: <\textit{paper title}>...;
 \ a$_{2}$: ...;  \ a$_{3}$: ...
\end{tcolorbox}
    \caption{The prompt example consists of three components: prefix\_question: a selective question; source\_node\_description: the description of the source node and its corresponding anchor nodes; candidate\_nodes\_description: the description of candidate neighbors and the anchor nodes corresponding to each candidate neighbor.}
    \label{fig:fig3}
\end{figure}

To enhance the capability of the large language models in distinguishing between various types of heterogeneous nodes, we additionally assign distinct type identifiers to the backend of each node. For example, a paper node could be described as "<p>[PA]". Following the formal definition provided above, Figure \ref{fig:fig3} illustrates a more intuitive prompt example for author disambiguation.

Our designed prompts do not explicitly capture the link information between nodes in the graph. Instead, we choose to describe key nodes in textual form based on their order of importance.
This decision arises from the complexity of inter-node connections, which often result in redundant contexts~\cite{fatemi2023talk}, making it challenging for large language models to comprehend. 
Consequently, there is a risk of LLMs diminishing the emphasis on node features, which are pivotal for our tasks. Nonetheless, the links among heterogeneous nodes remain crucial as they reflect their relationships and node significance. 
In following Sec.\ref{subsec:two-stage sampling}, we introduce a two-stage sampling approach to leverage structural information, prioritizing critical nodes and thereby enhancing the description of graph information.

\subsection{Two-Stage Sampling}\label{subsec:two-stage sampling}
In the previous subsection, we designed the unified prompt template for link prediction. However, as graph data becomes more complex, resembling the real world, employing a single prompt engineering approach becomes challenging in addressing practical application problems. 
Firstly, in large-scale graphs, attempting to describe the node information of $v$ using all $h$-hop neighbors, i.e. $k=\lvert\mathcal{V}_v^h\rvert$ formally, as shown in Eq.(\ref{eq:eq1}), leads to an uncontrollable prompt length.
Anothor issue arises due to substantial variations in the degrees of different node types. For example, the number of nodes in the $h$-hop subgraph around a paper node is significantly smaller than that around a field-type node. The two problems pose significant challenges to the input and contextual comprehension of LLMs.

In this work, we provide a two-stage sampling pipeline. The first stage aims to sample subgraphs $\mathcal{G}_{sub}^h(v)$ based on normalized degree from large-scale heterogeneous graphs while mitigating sampling bias caused by heterogeneous types. Subsequently, we obtain the top-$k$ anchor nodes sequence $\mathcal{AN}^h_k(v)$ through the second stage sampling with personalized pagerank to generate $d(v)$ in Eq.(\ref{eq:eq1}). The further details are as follows.

\noindent\textbf{Normalized Degree based Sampling} Inspired by previous studies ~\cite{hu2020heterogeneous, leskovec2006sampling}, we adopt a strategy for sampling heterogeneous subgraphs based on normalized degree. Specifically, this approach specifies the sampling probability of each hop's neighbors as their normalized degree. The normalized degree is defined as the node's degree normalized among all nodes of the same type in the same layer. Therefore, for the $l$-th layer subgraph sampling around central node $s$, the sampling probability of node $v$ can be described as follows:
\begin{equation}
    prob^{l}_s(v)=\frac{deg(v)^2}{\sum{deg(u)^2}}
\end{equation}
where node $u$ represents the neighbor node at the $l$-th layer within the subgraph, satisfying $u\in\mathcal{V}_s^h\setminus\mathcal{V}_s^{h-1}$ and $ \phi(v)=\phi(u)$.

The normalized degree based sampling in our first stage ensures that differences between node types are not ignored, preventing bias against certain node types (e.g., nodes with higher degrees are not indiscriminately considered more important). 
This approach maintains a similar number of different types of nodes in the subgraphs, thereby preserving richer semantic information. 
Furthermore, previous studies have demonstrated that leveraging up to 3-hop connectivity is effective for achieving excellent performance~\cite{kipf2016semi, velivckovic2017graph, hamilton2017inductive}. However, extending the information beyond 3-hop generally has a marginal impact on improvement and, in some cases, may even result in negative effects~\cite{cai2020note, zhang2021evaluating}. Therefore, we set the maximum value for multi-hop to 2-hop or 3-hop in our two-stage sampling approach.

\noindent\textbf{Sampling with Personalized PageRank} Through the sampling in the first stage, the heterogeneous subgraphs we obtain eliminate biases between different types, allowing all types of nodes to be compared regarding their importance on an equal footing. In the second stage, we directly compute the importance of all neighbor nodes within the subgraph $\mathcal{G}_{sub}^h(v)$ for the source node $s$ using Personalized PageRank (PPR)~\cite{bojchevski2020scaling, vattani2011preserving}. We then obtain the PPR vector $\vec{\pi_s}$ for the source node $s$ by iteratively updating the following:
\begin{equation}
    \vec{\pi_s}=\alpha * \vec{e_s} + (1-\alpha) * A^{\top} D^{-1}\vec{\pi_s}
\end{equation}
where $\alpha$ denotes the damping factor, $A$ stands for the adjacency matrix, $D^{-1}$ denotes the diagonal degree matrix and $\vec{\pi_s}$ signifies the unit vector.

This work employs a queue-based implementation of the equivalent random walk \cite{spitzer2013principles, wu2021unifying} to approximate PPR. Subsequently, the top-$k$ anchor nodes sequence $\mathcal{AN}^h_k(s)$ is obtained based on the ranking derived from PPR, which characterizes the top-$k$ neighbor nodes that are most critical for the source node $s$ within the whole hetero-graph.

The two-stage sampling restricts the generated link prediction prompt length to suit LLMs inputs while maximizing the retention of crucial neighborhood information pertaining to the target node within the subgraphs. It also makes use of the structural information on the graph, so that the generated anchor nodes $\mathcal{AN}^h_k(s)$ can be seen as a hub converting the graph structure into textual descriptions. This enables our prompts generated in Sec.\ref{subsec:prompt} to encompass not only node features but also implicit structural information.

\begin{table*}[th]
\centering
\renewcommand\arraystretch{1.2}
\resizebox{\textwidth}{!}{%
\begin{tabular}{c|cc|ccccc|ccccc} 
\toprule
Dataset & $\#$nodes & $\#$edges & $\#$papers & $\#$authors & $\#$fields & $\#$venues & $\#$institutes & $\#$P-A & $\#$P-F & $\#$P-V & $\#$A-I & $\#$P-P \\ 
\midrule
CS & 11,918,983 & 107,263,811 & 5,597,605 & 5,985,759 &  119,537&  27,433 & 16,931   & 15,571,614 & 47,462,559 & 5,597,606 & 7,190,480 & 31,441,552\\
\midrule
Mater & 4,552,941 & 42,161,581 & 2,442,235 & 2,005,362&  79,305 &  15,141&  10,898  &5,582,765 & 19,119,947&2,442,235 & 2,005,362& 13,011,272\\
\midrule
Engin & 5,191,920 & 36,146,719  & 3,239,504 & 1,819,100 &  99,444&  19,867&  14,005 & 3,741,135&  22,498,822&  3,239,504&  1,819,100 & 4,848,158\\
\midrule
Chem & 12,158,967 & 159,537,437  & 7,193,321 & 4,748,812 &  183,782&  19,142&  13,910 & 16,414,176&  57,162,528&  7,193,321&  4,748,812 & 74,018,600\\
\bottomrule
\end{tabular}%
}
\caption{OAG statistics.} 
\label{tab:stat} 
\end{table*}

\begin{figure}[t]
    \centering
    \includegraphics[width=0.50\textwidth]{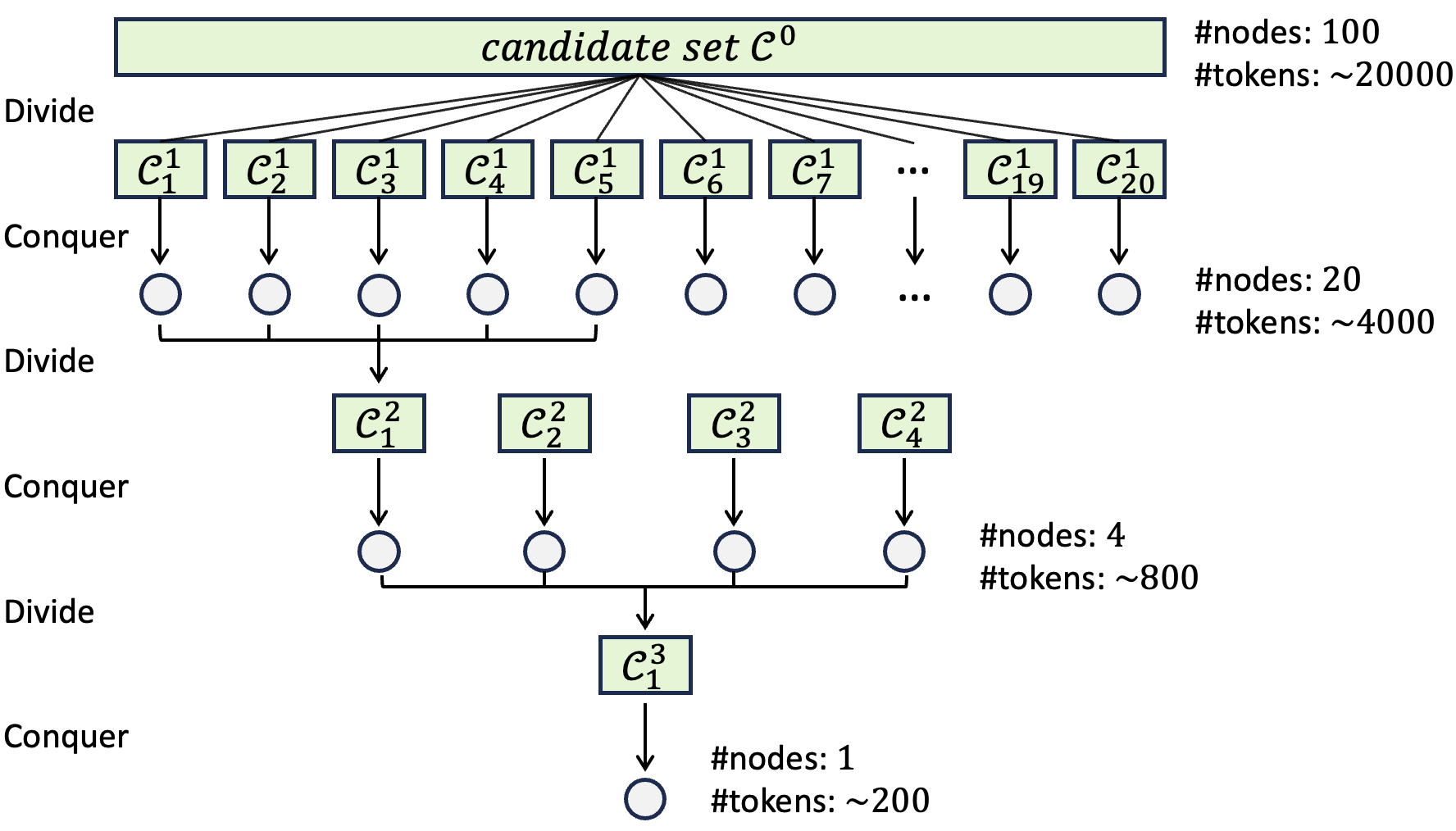}
    \caption{For a link prediction task involving 100 candidate neighbors, we set the candidate length limit $L$ to 5. The candidate neighbors can be divided into 20 sets, followed by three rounds of divide-and-conquer. This process ultimately yields a unique prediction result.}
    \label{fig:fig4}
\end{figure}

\subsection{Divide-and-Conquer Prediction} \label{subsec:Divide-and-conquer}
While the sampling pipeline addresses the potential issue of prompts length caused by Eq.(\ref{eq:eq1}), a careful observation of Eq.(\ref{eq:eq2}) reveals that an excessive number of candidate neighbors in link prediction, denoted as $\vert\mathcal{C}\vert$, also makes the prompt length uncontrollable. Especially in large-scale graphs, the high number of candidate neighbors poses a challenge in describing all of them within a single LLM's input window. For instance, in a link prediction task with 100 candidate neighbors, each node requires an average of approximately 200 tokens in the prompt for description. This results in a total of 20,000 tokens needed to describe all candidate neighbors, far exceeding the maximum token limit for a usual LLM's input window. Furthermore, the excessive number of candidate neighbors leads to redundant contexts, making it challenging for the LLMs to comprehend the input text.

LPNL avoids the aforementioned token overload by employing a divide-and-conquer strategy. Figure \ref{fig:fig4} provides an intuitive example of the divide-and-conquer prediction, allowing us to observe the descent of candidate neighbors and prompt tokens throughout the process. We set a length limit $L$ for the candidate set, ensuring that the length of all processed candidate sets does not exceed $L$. We represent $\mathcal{C}^i_j$ as the $j$-th candidate set of the $i$-th divide-and-conquer round. Specifically, for an original candidate set of length $\vert\mathcal{C}^0\vert$ where $\vert\mathcal{C}^0\vert > L$, we randomly divide it into $m$ sub-candidate sets, ensuring $m=\lceil\frac{\vert\mathcal{C}^{0}\vert}{L}\rceil$. This results in sets denoted as $\mathcal{C}_{1}^1, \mathcal{C}_{2}^1, ..., \mathcal{C}_m^1$, with the constraint that $max(\vert\mathcal{C}_{1}^1\vert, \vert\mathcal{C}_{2}^1\vert, ..., \vert\mathcal{C}_m^1\vert) \leq L$. 

As illustrated in Figure \ref{fig:fig1}, by employing the fine-tuned large language models to predict the candidate neighbor of the source node with the maximum link probability for each sub-candidate set, we can subsequently eliminate low-probability candidate neighbors. And the process generates new candidate sets based on the predicted results by refining the candidate sets. Specifically, for the candidate sets $C_{j+1}^i, C_{j+2}^i, ..., C_{j+k}^i$, a new candidate set $C_{k'}^{i+1}$ is generated in the following round based on their prediction results. The values of $k$ and $k'$ are determined based on the order of generation, ensuring that the condition $k \leq L$ is met. Following this divide-and-conquer process by refining candidate sets and making predictions, ultimately, we can obtain a unique prediction answer for the entire original candidate set $\mathcal{C}^0$.

\subsection{Self-Supervised Fine-tuning}
As a more relevant graph structure, large-scale graphs lack labelled data.
LPNL uses self-supervised learning for large language model fine-tuning. During the end-to-end prompt fine-tuning, it automatically constructs a candidate set containing ground truth, aligned with downstream prediction formats. The ground truth is used as the correct answer for link prediction. To ensure training correctness, the ground truth appears randomly within the candidate neighbor sequence. Notably, during the heterogeneous subgraph sampling process in Sec.\ref{subsec:two-stage sampling}, the edges between the ground truth and the source node are masked. Because the self-supervised fine-tuning does not require training labels provided by graph tasks, a fine-tuned LPNL model can make direct predictions on different graphs without the need of extra tuning.

\section{Experiments}
\subsection{Experiment Settings}
\begin{table*}[t]
\centering
\small
\renewcommand\arraystretch{1.5}
\setlength{\tabcolsep}{7pt}
\begin{tabular}{c||c||ccccc||c||l} 
\toprule
\midrule
{Dataset} & {Metric} & GraphSage  & HGT & RGCN & GCN  & GAT & LPNL & \ \ \ \ \ \ $\Delta$
\\ \midrule

    \multirow{3}{*}{\makecell{CS}} 
        ~  & NDCG &.814$\pm$.025 & .847$\pm$.042& .843$\pm$.056&.887$\pm$.031 & .911$\pm$.033 & \textbf{.985$\pm$.008} & $\uparrow 8.12\%$\\ ~& MRR &.640$\pm$.045&.712$\pm$.024 & .685$\pm$.056 &.727$\pm$.032 & .797$\pm$.051 &\textbf{.939$\pm$.018} &$\uparrow17.81\%$ \\~  & Hits@1 &.469$\pm$.012 & .562$\pm$.022& .532$\pm$.056&.568$\pm$.011 & .686$\pm$.014 & \textbf{.894$\pm$.004} & $\uparrow 30.32\%$
        \\\midrule
    \multirow{3}{*}{\makecell{Mater}} 
        ~  & NDCG &.765$\pm$.017 & .841$\pm$.034& .854$\pm$.042&.818$\pm$.016 & .897$\pm$.065 & \textbf{.954$\pm$.014} & $\uparrow6.35\%$\\ ~& MRR &.519$\pm$.052&.643$\pm$.031 & .665$\pm$.027 &.667$\pm$.036 & .747$\pm$.058 &\textbf{.881$\pm$.011} & $\uparrow17.93\%$ \\~  & Hits@1 &.278$\pm$.016 & .447$\pm$.019& .476$\pm$.028&.524$\pm$.032 & .597$\pm$.018 & \textbf{.809$\pm$.007} & $\uparrow35.51\%$
        \\\midrule
    \multirow{3}{*}{\makecell{Engin}} 
        ~  & NDCG &.798$\pm$.021 & .876$\pm$.022& .874$\pm$.061&.912$\pm$.041 & .913$\pm$.037 & \textbf{.977$\pm$.017} &$\uparrow7.01\%$ \\ ~& MRR &.570$\pm$.027&.691$\pm$.041 & .699$\pm$.034 &.747$\pm$.023 & .769$\pm$.041 &\textbf{.917$\pm$.017} &$\uparrow16.14\%$
        \\~  & Hits@1 &.342$\pm$.023 & .506$\pm$.018& .523$\pm$.056&.583$\pm$.021 & .624$\pm$.011 & \textbf{.858$\pm$.012} & $\uparrow37.50\%$ 
        \\\midrule
    \multirow{3}{*}{\makecell{Chem}} 
        ~  & NDCG &.821$\pm$.015 & .863$\pm$.015& .835$\pm$.036&.893$\pm$.017 & .899$\pm$.023 & \textbf{.955$\pm$.018} &$\uparrow 6.23\%$\\ ~& MRR &.649$\pm$.034&.724$\pm$.027 & .678$\pm$.031 &.749$\pm$.023 & .780$\pm$.029 &\textbf{.872$\pm$.038} &$\uparrow11.79\%$ \\~  & Hits@1 &.485$\pm$.024 & .523$\pm$.031& .530$\pm$.016&.609$\pm$.020 & .667$\pm$.022 & \textbf{.792$\pm$.007} &$\uparrow18.74\%$
        \\
        
\midrule
\bottomrule

\end{tabular}

\caption{Experimental results of different methods over the four datasets.} 
\label{tab:result} 
\end{table*}

\textbf{Models}
We fine-tune T5-base model~\cite{chung2022scaling} with a 1024 input window constraint as the backbone language model for our LPNL. The numbers of sampling hops $h=2$, top anchor nodes sequence $k=50$, and candidate length limit $L=3$ are used for all following experiments.
\\
\textbf{Datasets}
We conducted all experiments on the OAG, known as one of the largest publicly available heterogeneous graphs, comprising 178 million nodes and 2.236 billion edges. It includes five types of nodes (denoted as papers (P), authors (A), venues (V), institutes (I) and fields (F)) and their interrelations. In our specific experiments, we utilized four representative domain-specific subgraphs from OAG: Computer Science (CS), Material Science (Mater), Engineering (Engin) and Chemistry (Chem)~\cite{jiang2021pre}. The graph statistics are listed in Table \ref{tab:stat}. We partition each dataset into fine-tuning, validation, and test sets based on distinct time periods. Specifically, in the OAG dataset, papers are published between 1900 and 2019. Consequently, we utilize publications preceding 2015 for fine-tuning, data from 2015 to 2016 for validation, and information from 2016 onwards for testing.
\\
\textbf{Task}
We consider real-world link prediction tasks to evaluate the performance of our LPNL, specifically, author name disambiguation~\cite{ferreira2012brief}. Author name disambiguation is a fundamental challenge for curating academic publication and author information, as duplicated names are common. The objective is to predict the true author who has a genuine link with a given paper among all authors with the same name.
\\
\textbf{Baselines}
We select a series of supervised baselines, all of which are advanced graph neural network models. These include GCN~\cite{li2018deeper}, GraphSage~\cite{hamilton2017inductive} and GAT~\cite{velivckovic2017graph}, designed for homogeneous graphs, as well as RGCN~\cite{schlichtkrull2018modeling} and HGT~\cite{hu2020heterogeneous}, tailored for heterogeneous graphs.

\subsection{Overall Performance}

In this experiment, we compare the T5 model as the backbone version of LPNL to advanced GNN based baseline models across the four domain-specific subgraphs. We fine-tune the model separately across various subgraphs and evaluate the performance of the models in link prediction. The experimental results of the proposed method and baselines are summarized in Table \ref{tab:result}. All experiments for the author name disambiguation task over all datasets are evaluated in terms of NDCG, MRR and Hits@1~\cite{li2022learning, liu2009learning}.

The results show that in terms of all three metrics, the proposed LPNL significantly and consistently outperforms all baselines for all tasks on all datasets. Overall, our LPNL consistently yields the best performance among all methods, leading to an average improvement of 6.93\%, 15.92\% and 30.52\%, compared to the second best baseline method. Surprisingly, LPNL exhibited significant improvements in all settings, particularly in the Hit@1 metric. The substantial leap in achieving correct predictions with just a single attempt holds significant implications for practical applications. These improvements over GNNs indicate the efficacy of our proposed LPNL in enabling large language models to comprehend link prediction tasks within complex graphs and large language models have tremendous potential in addressing graph-related problems.

\begin{figure}[ht]
    \centering
    \includegraphics[width=0.5\textwidth]{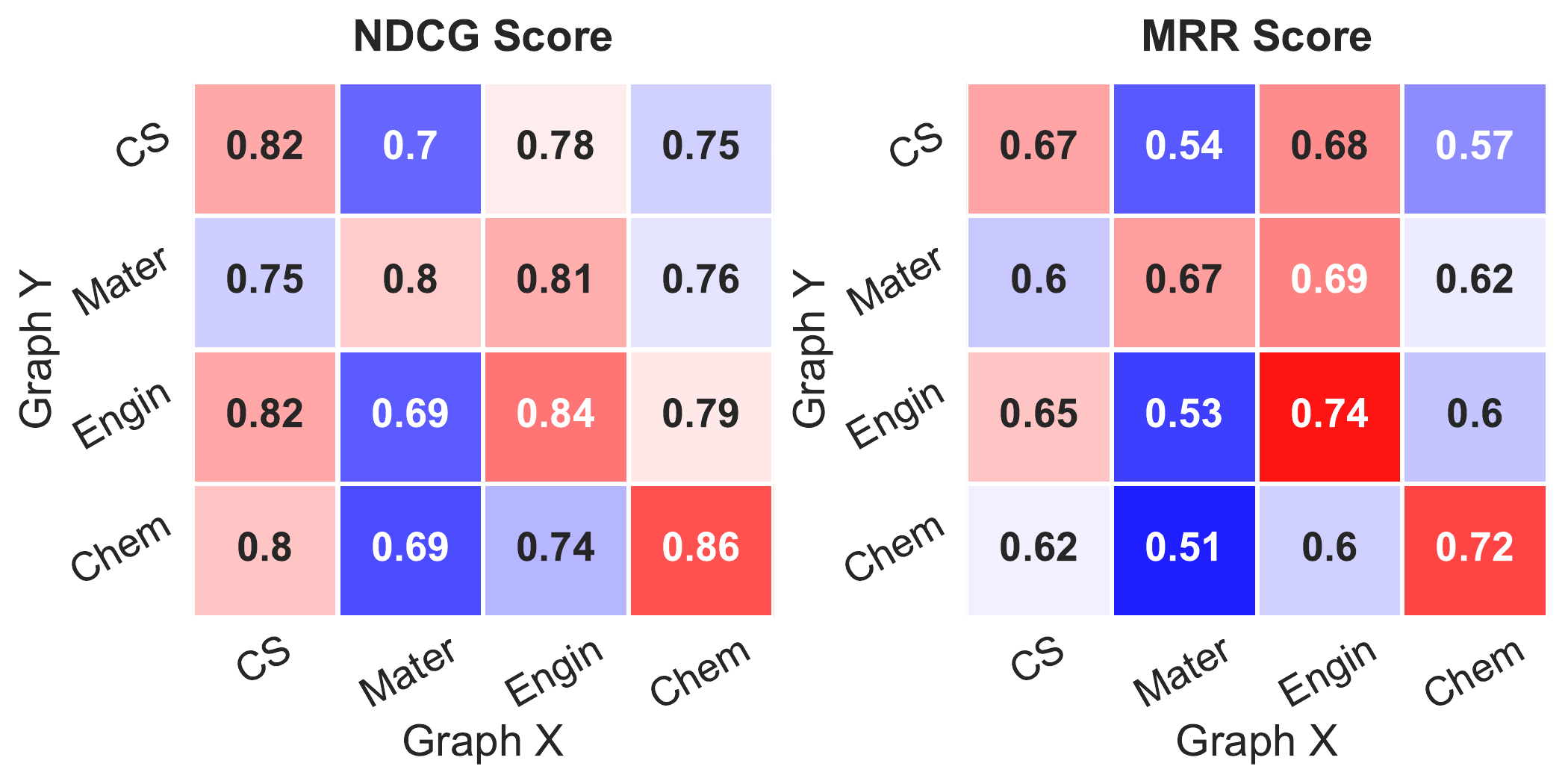}
    \caption{Cross-domain transfer results.}
    \label{fig:cross_domain}
\end{figure}

\subsection{Cross-Domain Knowledge Transfer}
To explore the generalization capabilities of LPNL, we set up experiments for cross-domain knowledge transfer. Specifically, we fine-tune the T5 model using LPNL on a graph corresponding to one domain and subsequently conducted testing on subgraphs from other domains. The experimental outcomes, visualized in Figure \ref{fig:cross_domain} as a heatmap, reveal that in most instances, the model exhibits optimal performance when fine-tuned within its original domain. Surprisingly, the models fine-tuned on other domains also demonstrate remarkably strong performance, often closely matching or even surpassing the best performance achieved by fine-tuning within the original domain (e.g., Mater-Engin). This highlights the robust knowledge transferability of our approach, which means it can make direct predictions on different graphs without the need for additional learning.

\begin{figure}[ht]
  \begin{subfigure}{4cm}
    \centering
    \includegraphics[width=1\linewidth]{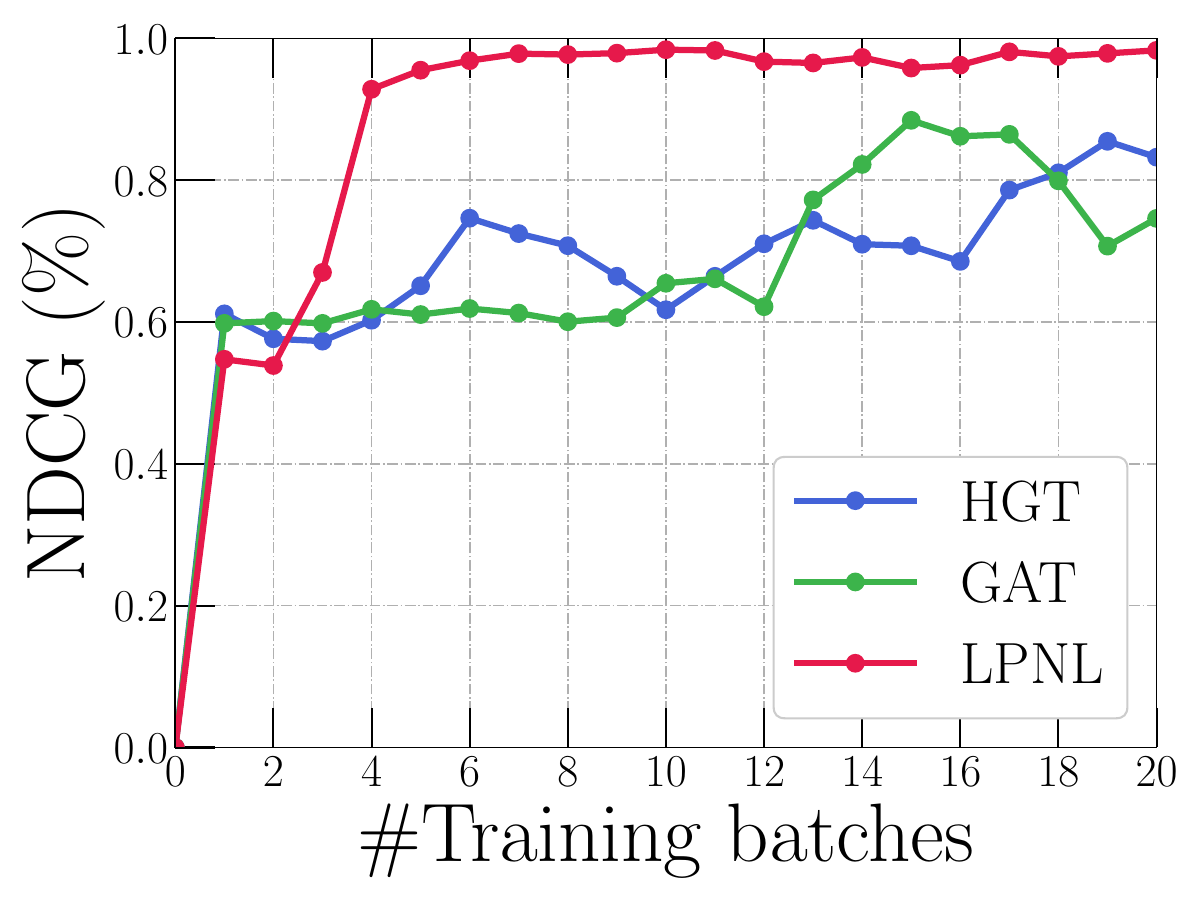}
    \label{fig:sub1}
  \end{subfigure}%
  \begin{subfigure}{4cm}
    \centering
    \includegraphics[width=1\linewidth]{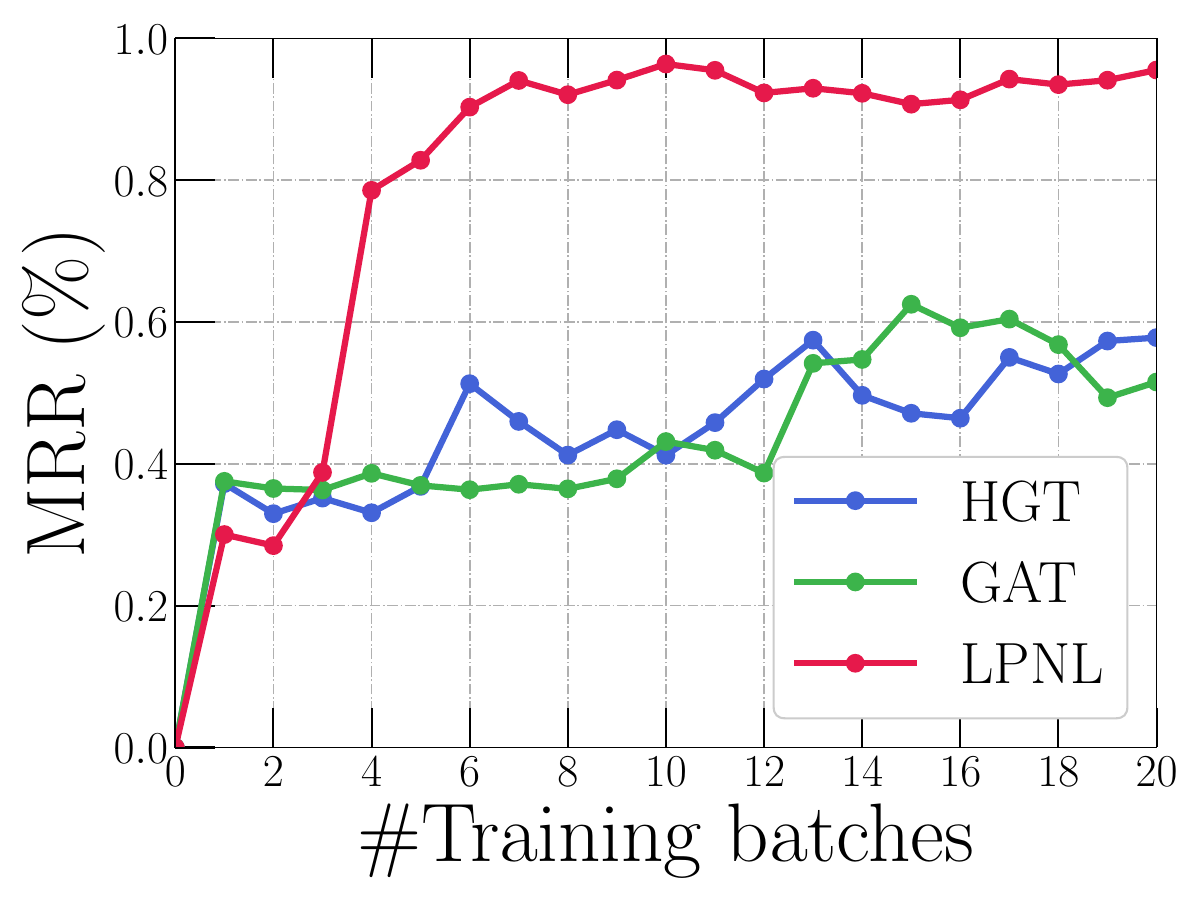}
    \label{fig:sub2}
  \end{subfigure}
  \vspace{-2.5em}
  \caption{LPNL converges fast in few-shot learning compared to GNNs.}
  \label{fig:few-shot}
\end{figure}

\subsection{Few-Shot Learning}
The extensive pretraining of large language models across various natural language tasks has endowed them with robust reasoning and generalization capabilities. In contrast to traditional GNN models, they require minimal training samples to converge and exhibit superior performance. We further investigates the few-shot learning capabilities of LPNL by comparing it with the top-performing homogeneous GNN and heterogeneous GNN in terms of overall performance. We configure the evaluate results to be printed every 1 batch, with each batch consisting of 50 link prediction tasks. We compare the few-shot results for the first 20 batches. The results in Table \ref{fig:few-shot} demonstrate that our LPNL swiftly converges with minimal sample fine-tuning, displaying comparable performance to the best fine-tuning outcomes. This showcases the portability of large language models in addressing graph-related tasks.



\begin{table}[ht]
\centering
\small
\renewcommand\arraystretch{1.1}
\setlength{\tabcolsep}{5pt}
\begin{tabular}{lccc} 
\toprule
 \textbf{Method} & \textbf{NDCG}  & \textbf{MRR} & \textbf{Hits@1}\\ \midrule 
 LPNL & \textbf{97.86} & \textbf{94.37} & \textbf{89.47} \\
 \midrule
 w/o Graph Info & 68.98 & 50.51 & 35.97 \\ 
 w/o Stage 1 & 76.31 & 57.29 & 41.83  \\ 
 w/o Stage 2 & 87.67 & 70.67 & 53.33  \\ 
\bottomrule
\end{tabular}
\caption{Ablation study results of sampling methods.} 
\label{tab:ablation1} 
\end{table}

\subsection{Ablation Study}

We conduct an ablation study on CS dataset to evaluate the effectiveness of our approach in employing large language models combined with graph knowledge strategies. We compare the performance among different versions of sampling methods: the standard LPNL, a version without any graph information, and another two sampling versions, each independently utilizing distinct stages. 
As illustrated in Table \ref{tab:ablation1}, the model's performance significantly diminishes when graph information is excluded. Furthermore, the performance of the versions without stage 1 and stage 2 shows a notable gap compared to LPNL. It 
indicates that employing our designed two-stage sampling pipeline enables LPNL to capture crucial information within the graph after balanced heterogeneous sampling, resulting in improved predictive outcomes.

\begin{table}[ht]
\centering
\small
\renewcommand\arraystretch{1.1}
\setlength{\tabcolsep}{5pt}
\begin{tabular}{cccc} 
\toprule
 \ \textbf{Hop} & \textbf{NDCG}  & \textbf{MRR} & \textbf{Hits@1}\\ \midrule 
 2-hop & \textbf{97.86} & \textbf{94.37} & \textbf{89.47} \\ 
 1-hop & 94.15 & 91.56 & 85.09 \\ 

\bottomrule
\end{tabular}
\caption{Ablation study results of multi-hop sampling.} 
\label{tab:ablation2} 
\end{table}

In our experiments, two critical operations contributing significantly to the outstanding performance of LPNL in link prediction are 2-hop and anchor nodes, which provide essential information to the LLMs. To assess the impact of these two key components on model performance, we conducted another ablation experiments, and the results are presented in Table \ref{tab:ablation2} and \ref{tab:ablation3}. It shows that incorporating multi-hop and more anchor nodes information can both enhance the LPNL's performance. However, further experiments indicate that increasing the number of hops and anchor nodes beyond a certain threshold does not lead to significant performance improvement. On the contrary, it may result in additional costs without notable benefits.

\begin{table}[ht]
\centering
\small
\renewcommand\arraystretch{1.1}
\setlength{\tabcolsep}{5pt}
\begin{tabular}{cccc} 
\toprule
 \ \textbf{\#Anchor Nodes} & \textbf{NDCG}  & \textbf{MRR} & \textbf{Hits@1}\\ \midrule 
 Top-30 & 92.86 & 76.48 & 62.05  \\ 
 Top-50 & 97.86 & \textbf{94.37} & \textbf{89.47} \\ 
  Top-70 & \textbf{98.06} & 93.84 & 88.69  \\ 
\bottomrule
\end{tabular}
\caption{Ablation study results of top-k anchor nodes} 
\label{tab:ablation3} 
\end{table}

\section{Related Work}

\noindent\textbf{Graph Representation Learning Based on GNNs}
Graph Neural Networks (GNNs) are the forefront of graph representation learning methods and have gained significant popularity across a range of graph-related tasks~\cite{wu2020comprehensive,zhou2020graph}. In these tasks, such as node classification and link prediction, GNNs-based approaches usually preprocess the corresponding text by a language model and encode the resulting embedding as node features. The final node representation is then obtained by aggregating the neighborhood features through spectral methods~\cite{bruna2013spectral,defferrard2016convolutional} and message passing~\cite{abu2019mixhop,hamilton2017inductive,schlichtkrull2018modeling}. Besides, some studies have attempted to propose the GNNs architectures on heterogeneous graphs~\cite{dong2020heterogeneous,wang2019heterogeneous, hu2020heterogeneous}. Notably, influenced by large language models, recent studies~\cite{sun2023all, huang2023prodigy}have explored the potential of GNNs in prompt learning. And there have also been attempts~\cite{ioannidis2022efficient, zhao2022learning} to explore collaborative training between Language Models and GNNs.

\noindent\textbf{Large Language Models with Graph Knowledge}
The emergence of large language models (LLMs) has propelled natural language processing (NLP) to new heights.~\cite{qiu2020pre}. For example, models like BERT~\cite{devlin2018bert} and T5~\cite{raffel2020exploring} demonstrate excellent performance in a wide range of downstream tasks, such as text classification and question answering. Besides, some works ~\cite{zhang2019ernie, liu2022oag, liu2020k}attempts to inject external graph knowledge into LLMs, thus enabling LLMs to gain the ability to solve problems on graphs. Recently, due to the powerful inferential capabilities of large language models, a burgeoning body of work~\cite{fatemi2023talk, ye2023natural, liu2023one} attempt to utilize natural language descriptions of graph features, employing generated prompts to instruct large language models in addressing various problems on graphs.

\section{Discussion}
From our experiments, we found that describing graphs using natural language does not follow the principle of "more information is better". Sampling more nodes can introduce additional information, but it may lead to information redundancy, resulting in a decline in the inferential capabilities of large language models. Therefore, the key lies in the setting of the sampling and divide-and-conquer length limits, which should align with the input window size and inferential capabilities of the large language models.
While designing prompts, we observe that complex relationships between nodes are challenging to articulate in text, especially in large or dense graphs, potentially leading to redundant contexts. LPNL leverages structural information during the sampling phase and, in the prompt generation, only conveys information about the sampled nodes. This approach aims to minimize context redundancy while maximizing the utilization of graph information.

\section{Conclusion}
In this paper, we explore, for the first time, the application of large language models to address the link prediction task on large-scale heterogeneous graphs. We introduce LPNL, a large language models based framework for scalable link prediction on large-scale graphs. We design specific prompt templates for the link prediction task and generate the prompts based on anchor nodes obtained through a two-stage sampling approach. These prompts are then input to the large language models for predictions. To tackle the token overload issue arising from an excessive number of candidate neighbors, we employ a divide-and-conquer strategy. Empirical evaluations demonstrate that LPNL achieves significant improvements compared to GNN baselines, showcasing its robust capability in cross-domain knowledge transfer and few-shot learning scenarios.

\section*{Limitations}
Some efforts in solving graph-related problems using LLMs involve supervised fine-tuning, resulting in limited ability for knowledge transfer. Although LPNL supports unsupervised learning without the need for labels, it is currently confined to link prediction tasks and has not been applied to a broader spectrum of graph-related tasks. LPNL has not yet explored larger parameter scales for large language models and their zero-shot potentials, which could provide increased input window sizes and enhanced inferential capabilities. Integrating our approach with other graph tasks and larger language models holds the potential to significantly improve predictive capabilities.

\section*{Ethics Statement}
Ethical considerations are of utmost importance in our research endeavors.  In this paper, we conscientiously adhere to ethical principles by exclusively utilizing open-source datasets and employing models that are either open-source or widely recognized in the scientific community.  Moreover, our proposed method is designed to ensure that the model does not produce any harmful or misleading information.  We are committed to upholding ethical standards throughout the research process, prioritizing transparency, and promoting the responsible use of technology for the betterment of society.


\bibliography{anthology,custom}

\begin{thebibliography}{42}
\expandafter\ifx\csname natexlab\endcsname\relax\def\natexlab#1{#1}\fi

\bibitem[{Abu-El-Haija et~al.(2019)Abu-El-Haija, Perozzi, Kapoor, Alipourfard, Lerman, Harutyunyan, Ver~Steeg, and Galstyan}]{abu2019mixhop}
Sami Abu-El-Haija, Bryan Perozzi, Amol Kapoor, Nazanin Alipourfard, Kristina Lerman, Hrayr Harutyunyan, Greg Ver~Steeg, and Aram Galstyan. 2019.
\newblock Mixhop: Higher-order graph convolutional architectures via sparsified neighborhood mixing.
\newblock In \emph{international conference on machine learning}, pages 21--29. PMLR.

\bibitem[{Bojchevski et~al.(2020)Bojchevski, Gasteiger, Perozzi, Kapoor, Blais, R{\'o}zemberczki, Lukasik, and G{\"u}nnemann}]{bojchevski2020scaling}
Aleksandar Bojchevski, Johannes Gasteiger, Bryan Perozzi, Amol Kapoor, Martin Blais, Benedek R{\'o}zemberczki, Michal Lukasik, and Stephan G{\"u}nnemann. 2020.
\newblock Scaling graph neural networks with approximate pagerank.
\newblock In \emph{Proceedings of the 26th ACM SIGKDD International Conference on Knowledge Discovery \& Data Mining}, pages 2464--2473.

\bibitem[{Bruna et~al.(2013)Bruna, Zaremba, Szlam, and LeCun}]{bruna2013spectral}
Joan Bruna, Wojciech Zaremba, Arthur Szlam, and Yann LeCun. 2013.
\newblock Spectral networks and locally connected networks on graphs.
\newblock \emph{arXiv preprint arXiv:1312.6203}.

\bibitem[{Cai and Wang(2020)}]{cai2020note}
Chen Cai and Yusu Wang. 2020.
\newblock A note on over-smoothing for graph neural networks.
\newblock \emph{arXiv preprint arXiv:2006.13318}.

\bibitem[{Cai et~al.(2021)Cai, Li, Wang, and Ji}]{cai2021line}
Lei Cai, Jundong Li, Jie Wang, and Shuiwang Ji. 2021.
\newblock Line graph neural networks for link prediction.
\newblock \emph{IEEE Transactions on Pattern Analysis and Machine Intelligence}, 44(9):5103--5113.

\bibitem[{Chung et~al.(2022)Chung, Hou, Longpre, Zoph, Tay, Fedus, Li, Wang, Dehghani, Brahma et~al.}]{chung2022scaling}
Hyung~Won Chung, Le~Hou, Shayne Longpre, Barret Zoph, Yi~Tay, William Fedus, Yunxuan Li, Xuezhi Wang, Mostafa Dehghani, Siddhartha Brahma, et~al. 2022.
\newblock Scaling instruction-finetuned language models.
\newblock \emph{arXiv preprint arXiv:2210.11416}.

\bibitem[{Defferrard et~al.(2016)Defferrard, Bresson, and Vandergheynst}]{defferrard2016convolutional}
Micha{\"e}l Defferrard, Xavier Bresson, and Pierre Vandergheynst. 2016.
\newblock Convolutional neural networks on graphs with fast localized spectral filtering.
\newblock \emph{Advances in neural information processing systems}, 29.

\bibitem[{Devlin et~al.(2018)Devlin, Chang, Lee, and Toutanova}]{devlin2018bert}
Jacob Devlin, Ming-Wei Chang, Kenton Lee, and Kristina Toutanova. 2018.
\newblock Bert: Pre-training of deep bidirectional transformers for language understanding.
\newblock \emph{arXiv preprint arXiv:1810.04805}.

\bibitem[{Dong et~al.(2020)Dong, Hu, Wang, Sun, and Tang}]{dong2020heterogeneous}
Yuxiao Dong, Ziniu Hu, Kuansan Wang, Yizhou Sun, and Jie Tang. 2020.
\newblock Heterogeneous network representation learning.
\newblock In \emph{IJCAI}, volume~20, pages 4861--4867.

\bibitem[{Fatemi et~al.(2023)Fatemi, Halcrow, and Perozzi}]{fatemi2023talk}
Bahare Fatemi, Jonathan Halcrow, and Bryan Perozzi. 2023.
\newblock Talk like a graph: Encoding graphs for large language models.
\newblock \emph{arXiv preprint arXiv:2310.04560}.

\bibitem[{Ferreira et~al.(2012)Ferreira, Gon{\c{c}}alves, and Laender}]{ferreira2012brief}
Anderson~A Ferreira, Marcos~Andr{\'e} Gon{\c{c}}alves, and Alberto~HF Laender. 2012.
\newblock A brief survey of automatic methods for author name disambiguation.
\newblock \emph{Acm Sigmod Record}, 41(2):15--26.

\bibitem[{Hamilton et~al.(2017)Hamilton, Ying, and Leskovec}]{hamilton2017inductive}
Will Hamilton, Zhitao Ying, and Jure Leskovec. 2017.
\newblock Inductive representation learning on large graphs.
\newblock \emph{Advances in neural information processing systems}, 30.

\bibitem[{Hu et~al.(2020)Hu, Dong, Wang, and Sun}]{hu2020heterogeneous}
Ziniu Hu, Yuxiao Dong, Kuansan Wang, and Yizhou Sun. 2020.
\newblock Heterogeneous graph transformer.
\newblock In \emph{Proceedings of the web conference 2020}, pages 2704--2710.

\bibitem[{Huang et~al.(2020)Huang, Wang, and Wang}]{huang2020analysis}
Han Huang, Hongyu Wang, and Xiaoguang Wang. 2020.
\newblock An analysis framework of research frontiers based on the large-scale open academic graph.
\newblock \emph{Proceedings of the Association for Information Science and Technology}, 57(1):e307.

\bibitem[{Huang et~al.(2023)Huang, Ren, Chen, Kr{\v{z}}manc, Zeng, Liang, and Leskovec}]{huang2023prodigy}
Qian Huang, Hongyu Ren, Peng Chen, Gregor Kr{\v{z}}manc, Daniel Zeng, Percy Liang, and Jure Leskovec. 2023.
\newblock Prodigy: Enabling in-context learning over graphs.
\newblock \emph{arXiv preprint arXiv:2305.12600}.

\bibitem[{Ioannidis et~al.(2022)Ioannidis, Song, Zheng, Zhang, Ma, Xu, Zeng, Chilimbi, and Karypis}]{ioannidis2022efficient}
Vassilis~N Ioannidis, Xiang Song, Da~Zheng, Houyu Zhang, Jun Ma, Yi~Xu, Belinda Zeng, Trishul Chilimbi, and George Karypis. 2022.
\newblock Efficient and effective training of language and graph neural network models.
\newblock \emph{arXiv preprint arXiv:2206.10781}.

\bibitem[{Jiang et~al.(2021)Jiang, Jia, Fang, Shi, Lin, and Wang}]{jiang2021pre}
Xunqiang Jiang, Tianrui Jia, Yuan Fang, Chuan Shi, Zhe Lin, and Hui Wang. 2021.
\newblock Pre-training on large-scale heterogeneous graph.
\newblock In \emph{Proceedings of the 27th ACM SIGKDD conference on knowledge discovery \& data mining}, pages 756--766.

\bibitem[{Kipf and Welling(2016)}]{kipf2016semi}
Thomas~N Kipf and Max Welling. 2016.
\newblock Semi-supervised classification with graph convolutional networks.
\newblock \emph{arXiv preprint arXiv:1609.02907}.

\bibitem[{Leskovec and Faloutsos(2006)}]{leskovec2006sampling}
Jure Leskovec and Christos Faloutsos. 2006.
\newblock Sampling from large graphs.
\newblock In \emph{Proceedings of the 12th ACM SIGKDD international conference on Knowledge discovery and data mining}, pages 631--636.

\bibitem[{Li(2022)}]{li2022learning}
Hang Li. 2022.
\newblock \emph{Learning to rank for information retrieval and natural language processing}.
\newblock Springer Nature.

\bibitem[{Li et~al.(2018)Li, Han, and Wu}]{li2018deeper}
Qimai Li, Zhichao Han, and Xiao-Ming Wu. 2018.
\newblock Deeper insights into graph convolutional networks for semi-supervised learning.
\newblock In \emph{Proceedings of the AAAI conference on artificial intelligence}, volume~32.

\bibitem[{Liu et~al.(2023)Liu, Feng, Kong, Liang, Tao, Chen, and Zhang}]{liu2023one}
Hao Liu, Jiarui Feng, Lecheng Kong, Ningyue Liang, Dacheng Tao, Yixin Chen, and Muhan Zhang. 2023.
\newblock One for all: Towards training one graph model for all classification tasks.
\newblock \emph{arXiv preprint arXiv:2310.00149}.

\bibitem[{Liu et~al.(2009)}]{liu2009learning}
Tie-Yan Liu et~al. 2009.
\newblock Learning to rank for information retrieval.
\newblock \emph{Foundations and Trends{\textregistered} in Information Retrieval}, 3(3):225--331.

\bibitem[{Liu et~al.(2020)Liu, Zhou, Zhao, Wang, Ju, Deng, and Wang}]{liu2020k}
Weijie Liu, Peng Zhou, Zhe Zhao, Zhiruo Wang, Qi~Ju, Haotang Deng, and Ping Wang. 2020.
\newblock K-bert: Enabling language representation with knowledge graph.
\newblock In \emph{Proceedings of the AAAI Conference on Artificial Intelligence}, volume~34, pages 2901--2908.

\bibitem[{Liu et~al.(2022)Liu, Yin, Zheng, Zhang, Zhang, Yang, Dong, and Tang}]{liu2022oag}
Xiao Liu, Da~Yin, Jingnan Zheng, Xingjian Zhang, Peng Zhang, Hongxia Yang, Yuxiao Dong, and Jie Tang. 2022.
\newblock Oag-bert: Towards a unified backbone language model for academic knowledge services.
\newblock In \emph{Proceedings of the 28th ACM SIGKDD Conference on Knowledge Discovery and Data Mining}, pages 3418--3428.

\bibitem[{Qiu et~al.(2020)Qiu, Sun, Xu, Shao, Dai, and Huang}]{qiu2020pre}
Xipeng Qiu, Tianxiang Sun, Yige Xu, Yunfan Shao, Ning Dai, and Xuanjing Huang. 2020.
\newblock Pre-trained models for natural language processing: A survey.
\newblock \emph{Science China Technological Sciences}, 63(10):1872--1897.

\bibitem[{Raffel et~al.(2020)Raffel, Shazeer, Roberts, Lee, Narang, Matena, Zhou, Li, and Liu}]{raffel2020exploring}
Colin Raffel, Noam Shazeer, Adam Roberts, Katherine Lee, Sharan Narang, Michael Matena, Yanqi Zhou, Wei Li, and Peter~J Liu. 2020.
\newblock Exploring the limits of transfer learning with a unified text-to-text transformer.
\newblock \emph{The Journal of Machine Learning Research}, 21(1):5485--5551.

\bibitem[{Schlichtkrull et~al.(2018)Schlichtkrull, Kipf, Bloem, Van Den~Berg, Titov, and Welling}]{schlichtkrull2018modeling}
Michael Schlichtkrull, Thomas~N Kipf, Peter Bloem, Rianne Van Den~Berg, Ivan Titov, and Max Welling. 2018.
\newblock Modeling relational data with graph convolutional networks.
\newblock In \emph{The Semantic Web: 15th International Conference, ESWC 2018, Heraklion, Crete, Greece, June 3--7, 2018, Proceedings 15}, pages 593--607. Springer.

\bibitem[{Shi et~al.(2016)Shi, Li, Zhang, Sun, and Philip}]{shi2016survey}
Chuan Shi, Yitong Li, Jiawei Zhang, Yizhou Sun, and S~Yu Philip. 2016.
\newblock A survey of heterogeneous information network analysis.
\newblock \emph{IEEE Transactions on Knowledge and Data Engineering}, 29(1):17--37.

\bibitem[{Spitzer(2013)}]{spitzer2013principles}
Frank Spitzer. 2013.
\newblock \emph{Principles of random walk}, volume~34.
\newblock Springer Science \& Business Media.

\bibitem[{Sun et~al.(2023)Sun, Cheng, Li, Liu, and Guan}]{sun2023all}
Xiangguo Sun, Hong Cheng, Jia Li, Bo~Liu, and Jihong Guan. 2023.
\newblock All in one: Multi-task prompting for graph neural networks.

\bibitem[{Vattani et~al.(2011)Vattani, Chakrabarti, and Gurevich}]{vattani2011preserving}
Andrea Vattani, Deepayan Chakrabarti, and Maxim Gurevich. 2011.
\newblock Preserving personalized pagerank in subgraphs.
\newblock In \emph{ICML}, volume~11, pages 793--800.

\bibitem[{Veli{\v{c}}kovi{\'c} et~al.(2017)Veli{\v{c}}kovi{\'c}, Cucurull, Casanova, Romero, Lio, and Bengio}]{velivckovic2017graph}
Petar Veli{\v{c}}kovi{\'c}, Guillem Cucurull, Arantxa Casanova, Adriana Romero, Pietro Lio, and Yoshua Bengio. 2017.
\newblock Graph attention networks.
\newblock \emph{arXiv preprint arXiv:1710.10903}.

\bibitem[{Wang et~al.(2019)Wang, Ji, Shi, Wang, Ye, Cui, and Yu}]{wang2019heterogeneous}
Xiao Wang, Houye Ji, Chuan Shi, Bai Wang, Yanfang Ye, Peng Cui, and Philip~S Yu. 2019.
\newblock Heterogeneous graph attention network.
\newblock In \emph{The world wide web conference}, pages 2022--2032.

\bibitem[{Wu et~al.(2021)Wu, Gan, Wei, and Zhang}]{wu2021unifying}
Hao Wu, Junhao Gan, Zhewei Wei, and Rui Zhang. 2021.
\newblock Unifying the global and local approaches: an efficient power iteration with forward push.
\newblock In \emph{Proceedings of the 2021 International Conference on Management of Data}, pages 1996--2008.

\bibitem[{Wu et~al.(2020)Wu, Pan, Chen, Long, Zhang, and Philip}]{wu2020comprehensive}
Zonghan Wu, Shirui Pan, Fengwen Chen, Guodong Long, Chengqi Zhang, and S~Yu Philip. 2020.
\newblock A comprehensive survey on graph neural networks.
\newblock \emph{IEEE transactions on neural networks and learning systems}, 32(1):4--24.

\bibitem[{Ye et~al.(2023)Ye, Zhang, Wang, Xu, and Zhang}]{ye2023natural}
Ruosong Ye, Caiqi Zhang, Runhui Wang, Shuyuan Xu, and Yongfeng Zhang. 2023.
\newblock Natural language is all a graph needs.
\newblock \emph{arXiv preprint arXiv:2308.07134}.

\bibitem[{Zhang and Chen(2018)}]{zhang2018link}
Muhan Zhang and Yixin Chen. 2018.
\newblock Link prediction based on graph neural networks.
\newblock \emph{Advances in neural information processing systems}, 31.

\bibitem[{Zhang et~al.(2021)Zhang, Sheng, Jiang, Xia, Gao, Yang, and Cui}]{zhang2021evaluating}
Wentao Zhang, Zeang Sheng, Yuezihan Jiang, Yikuan Xia, Jun Gao, Zhi Yang, and Bin Cui. 2021.
\newblock Evaluating deep graph neural networks.
\newblock \emph{arXiv preprint arXiv:2108.00955}.

\bibitem[{Zhang et~al.(2019)Zhang, Han, Liu, Jiang, Sun, and Liu}]{zhang2019ernie}
Zhengyan Zhang, Xu~Han, Zhiyuan Liu, Xin Jiang, Maosong Sun, and Qun Liu. 2019.
\newblock Ernie: Enhanced language representation with informative entities.
\newblock \emph{arXiv preprint arXiv:1905.07129}.

\bibitem[{Zhao et~al.(2022)Zhao, Qu, Li, Yan, Liu, Li, Xie, and Tang}]{zhao2022learning}
Jianan Zhao, Meng Qu, Chaozhuo Li, Hao Yan, Qian Liu, Rui Li, Xing Xie, and Jian Tang. 2022.
\newblock Learning on large-scale text-attributed graphs via variational inference.
\newblock \emph{arXiv preprint arXiv:2210.14709}.

\bibitem[{Zhou et~al.(2020)Zhou, Cui, Hu, Zhang, Yang, Liu, Wang, Li, and Sun}]{zhou2020graph}
Jie Zhou, Ganqu Cui, Shengding Hu, Zhengyan Zhang, Cheng Yang, Zhiyuan Liu, Lifeng Wang, Changcheng Li, and Maosong Sun. 2020.
\newblock Graph neural networks: A review of methods and applications.
\newblock \emph{AI open}, 1:57--81.

\end{thebibliography}
\bibliographystyle{acl_natbib}




\end{document}